# To Reduce Gross NPA and Classify Defaulters Using Shannon Entropy

Nikhil Sonavane[1] , Ambarish Moharil[2], Chirag Kedia[3],  Mansimran Singh Anand[4]

[1] Vishwakarma Institute of Technology, Pune
[2] Vishwakarma Institute of Technology, Pune
[3] Vellore Institute of Technology, Vellore
[4] Vellore Institute of Technology, Vellore

Non Performing Asset(NPA) has been in a serious attention by banks over the past few years. NPA cause a huge loss to the banks, hence it becomes an extremely critical step in deciding which loans have the capabilities to become an NPA, and thereby deciding which loans to grant and which ones to reject. In this paper, which focuses on the exact crux of the matter, we have proposed an algorithm which is designed to handle the financial data very meticulously to predict with a very high accuracy whether a particular loan would be classified as a NPA in future or not.  Instead of the conventional, less accurate classifiers used to decide which loans can turn to be NPA, we build our own classifier model using Entropy as the base. We have created an entropy based classifier using Shannon Entropy. The classifier model categorizes our data points in two categories, 'accepted' or 'rejected'. We make use of local entropy and global entropy to help us determine the output. The entropy classifier model is then compared with existing classifiers used to predict NPAs, thereby giving us an idea about the performance.

# 1 Introduction

A healthy and efficient banking system is integral to the economic success of any country. The Government of India and RBI take notable and substantial steps time to time to formulate regulation, steps and policies that need to be taken to make this section stronger. These initiatives have been taken to make sure that the stability and efficiency of the bank are not compromised. Even though promising and stricter rules have been passed the world has been hit by numerous Financial crisis[1]. The Global financial crisis of 2008 was an eye opener for many. These crises made people realize that failures in the banking sector can have inauspicious effects on economic activity more than another business sector of the country.

[6]Among many, Asset quality has emerged as one of the primitive indicators of a healthy banking system, as they help to provide understanding on the stability and firmness of the financial system. One of the major threats to the stability of the banking system is the depletion of its asset quality that would end up as a Non – performing Asset (NPA). [2]According to RBI (2015a), *"An asset becomes NPA if, interest and/or installment of principal remain overdue for a period of more than 90 days."* The growth of NPA in terms of its value majorly affects the growth and productivity of the banks. NPA can be further classified into 3 categories, Substandard NPA, doubtful NPA and Loss asset[5].

   a. Substandard NPA – An asset is classified as a Substandard NPA if it has remained an NPA for a time span of less than or equal to 12 months. Such an asset is defined as a "credit weakness" and hence is characterized with by a distinct possibility that the bank will incur a loss if the above inadequacy is not fixed.

   b. Doubtful NPA – An asset is categorized as a doubtful NPA if it has remained in the Substandard category for a period of 12 months. Based on the current know facts and conditions the value of the asset is highly questionable and doubtful once it is classified as a doubtful NPA.

   c. Loss Asset – An asset is classified as a loss asset if it has been recognized as a loss by the bank or the auditors but the asset amount hasn't been written off wholly.

[8][4]The gross NPA as a percentage of loans as of March 2019 stands at 9.3%, this number predicted to increase in the coming years. Various efforts and steps have been taken to inhibit and restrain the growth of NPA. Factors leading to growth of NPA can range from mismanagement at the staff level, inappropriate lending rules of the bank, deviation of funds, growing number of defaults and economic conditions of the country. Despite the measure taken the issue and drawback faced by the bank in regards to NPA are on the rise. We believe that the growing percentages of NPA can be brought down if we can precisely predict the factors contributing to the occurrence of NPA.

This can be achieved by accurately predicting and visualizing the data related to NPA. Here while working on the prediction it is observed that the analysis should not be restricted to that NPA but also the factors affecting it should be precisely monitored. Hence, we believe that the analysis of such sort of data will require a different perspective as compared to that of the prediction of any other dataset. Analytical reasoning is cardinal to the task of the analyst when it comes to applying human cognitive judgement to reach a conclusion keeping in mind the assumptions and evidences. The job of financial data analysis is considered one of the most challenging tasks for any data analysis.

In this paper we have proposed an algorithm which is designed to handle the financial data very meticulously to predict with a very high accuracy whether a particular loan would be classified as a NPA in future or not. The rest of the paper is organized as follows: Section 2, Entropy – A brief overview on the concept used to classify the data; Section 3, Methodology – The exact steps used by the algorithm right from collection of data to showcasing of the results; Section 4, Architecture – A flow chart of various steps our algorithm takes; Section 5, Dataset – A brief overview of the dataset used; Section 6 and 7 – Result and Conclusions.

## 2  Entropy

The main idea of this paper revolves around the central concept of entropy. Entropy is the quantity that denotes the total randomness or uncertainty of a variable. If the sample of data that is processed is completely homogenous then the value of entropy will be zero (as there is no randomness) on the other hand if the sample of data is equally divided then the value of entropy for that dataset will be one[3].

[3]Using the above-mentioned ideology, we have worked our way around this paper. The type of entropy used in our paper is Shannon entropy which was formulated by Claude E. Shannon. Shannon Entropy tells the uncertainty associated with a variable, hence allowing us to denote the average number of bits needed to encode a string of symbols, depending on their frequency.

Given a set of values v € V, the entropy H(V) is denoted by the following equation,

$$H(V) = - \sum_{v \in V} P(v) log_2[P(v)]$$

(1)

In this paper, we have used metrics to detect "anomalous" values based on the entropy difference we get when the instance is added to both the positive and negative dataset. The bigger anomaly we get based on comparison of entropy values will help us decide whether this new instance will be classified as NPA or not.

## 3  Methodology

Our approach towards solving this major problem was state of the art and unique and as described by the results in section (Results Section), they form a strong argument regarding its uniqueness and novelty. Classification problems are generally observed to face problems due to unorganized data structures. The reason a lot of these models fail is due to the data bias present while training the model. [3]The number of data points belonging to either of the classes is a very important factor as it determines the skewness of the data and predicts how the model is going to learn. If the number of positive cases is greater than the negative cases in a significant amount, then the probability of the model overfitting on the positive cases is more likely. And at times, it becomes very difficult to analyze the amount of overfitting the model has undergone in presence of a huge dataset[3]. In our dataset as explained in section 5, we had 4,06,601 negative or defaulter data points and 1,25,827 positive or non-defaulter data points. As it can be clearly observed that the dataset used in this experiment was extremely skewed and would have probably resulted into over-fitting of the model.

[3][4]To solve this problem, we deduced a very quirky method. After pre-processing of the data, we moved onto calculating the entropy of a single attribute. We moved on from local entropy calculation (μ) to global entropy calculation (Ω) and finally calculation of the difference between a reference metric of these entropies (Difference of Entropy Metrics or DEM) which decides whether an entity or a data point belongs to a particular class.

The following points describe our approach and methodology: -

Divide: - [3]The collected dataset was analyzed for the number of data points belonging to either of the classes. As stated above, the data was observed to have 4,06,601 defaulter data points and 1,25,827 non-defaulter data points. Sampling this data with equal number of data points would have been resulted into the loss of data, so it was decided to avoid this procedure. Instead of sampling the data, we applied the simple approach of dividing the data into two respective classes. Two classes C+ and C- were generated consisting the non-defaulter data and the defaulter data respectively. With this, we eliminated the bias present in the dataset. In class C+ as all cases were of non-defaulters so, the bias was completely eliminated. Also, if we look at the output column of "loan_status" the probability of occurrence was same for each and every case out of the 1,25,827 cases of non-defaulters and same for the 4,06,601 Defaulter cases. The advantage we get out of this approach is not just eliminating the bias, but the occurrence probability of each data point. This occurrence probability forms a very basic and foundational assumption while calculating the entropy of each attribute. In thermodynamics, while calculating the Boltzmann Entropy for a certain gas, a primal assumption is made that all the gas molecules of a certain gas exist in the same state of randomness. We extend this assumption to our model and assume that all the data points belonging to a class have same probability of occurrence and are present in the same state of randomness. So, all the defaulter cases have same probability of occurrence in the C- class and all the non-defaulter cases have the same probability of occurrence in the C+ class. This division of data makes sense while calculating the respective entropy of the C+ and the C- classes.

Local Entropy Calculation (µ): - Local Entropy (µ) is calculated using the Shannon Entropy. In the C+ and C- sets we have 88 total attributes. Each attribute is considered as an independent entity. And the probability of its occurrence is taken as mutually independent from the others. We calculate and plot µ along with the attributes which are assumed as independent variables. Shannon Entropy is given by the following formula: -

$$H(X) = -\sum_{i=1}^{n} p(x_i) \log_b p(x_i)$$

(2)

To calculate the local entropy, we calculate the Shannon Entropy of each attribute as µ1, µ2…....... µn. Here the entropy of each attribute is assumed to be independent of the other as we consider them as mutually independent events while calculating the respective probabilities. [3][7]Shannon Entropy gives the information of randomness of a variable, Shannon Entropy of the dataset attribute is assumed to be the measure of uncertainty present in the respective attribute given that it belongs to either of the classes. So µ1 subset C+ gives the randomness associated with the 1st attribute of the non-defaulter class. Now, as stated regarding the assumption of these events being mutually exclusive or independent, we calculate the local sum. Local sum can be defined as the sum of entropies of each attribute belonging to a respective class of data. So basically, we calculate two local sums. One belonging to the C+ non-defaulter class and the other belonging to the C- defaulter class.

Local Sum (α) = µ1+ µ2.......... +µn          (3)

This local sum is further referred as reference entropy metric. Reference Entropy Metric gives us a number or a metric for comparison of local and global entropy. It is further used to calculate the difference between the local and the global entropy that decides whether a case belongs to the defaulter class or a non-defaulter class.

```
loan_amnt
 0.9978900487998066
int_rate
 0.9998373720025702
pymnt_plan
 0.00025810525994835793
dti
 0.9987989291622914
delinq_2yrs
 1.0466575534958085
mths_since_last_delinq
 0.8479136005381216
mths_since_last_record
 -0.0
open_acc
 0.9899996730562759
```

Fig: 1 – Local Entropy Values of different Attributes

Global Entropy Calculation (Ω): - Global Entropy (Ω) is calculated by inserting a new data point into the existing data pool. Using the global entropy or Ω we calculate or observe the change in the overall attribute entropy of the class set. Suppose we have a collection set T, where T = {t1, t2, …....tn} and t1, t2, …...tn are random set of data points to be classified. So, in order to classify the data points, we insert them one by one into separated data pools and calculate if there is any change T subset C+ or T subset C-. This whole model works on the physical sense of change in entropy. To present an overview, consider an example of determining whether an ion is positively or negatively charged. If the ion is set into a pool of protons and shows no change in the existing entropy or shows slight change then this ion might be positively charged. It can be verified by dropping the ion into a pool of negatively charged electrons. If the Δ or change in entropy of the ion is high as well as there is a change in the global entropy of the electron pool, we deduce that the ion is positively charged[3][4][7]. Similarly, when an unknown data point is put in an observation set of C+ and C- classes, we observe the change in the entropy of the entire class and the entropy it gains after the data point is added is called as Global Entropy Ω. Further in the same way as local entropy metric is calculated, we calculate the global entropy metric by taking the sum of individual attribute entropies.

Global Sum (β) = Ω1+ Ω2+….......... Ωn                                                                           (4)

This global sum is further referred as Final Entropy Metric β.

Difference of Entropy Metrics (DEM): - DEM is defined as the difference between Reference Entropy Metric α and Final Entropy Metric β.

DEM = α - β                                                                                                                      (5)

DEM gives the result whether the data point is a defaulter case or a non-defaulter case. If DEM is positive or the change from local entropy to global entropy is reduced, it means that the data point belongs to that class. If the change is negative, indicating that β > α points then the data belongs to the other class. When a positive case is inserted in a positive data pool, the global entropy will increase slightly but if it is inserted into a negative data pool the change in entropy is significant and vice versa. We classify the data points based on this change or Difference of Entropy Metrics (Reference Entropy Metric – Final Entropy Metric). This approach is found to be accurate as compared to other models because it incorporates a physical phenomenon and is based on universally defined truths or axioms[3].

```
In [200]:  1 entropy_predict(pos_sample, data_n1) #Positive Sample to Negative Data Pool
           Reference_Entropy 50.68159246809343
           Final_Entropy 50.683911098891734
Out[200]: -0.0023186307983067422

In [201]:  1 entropy_predict(pos_sample, data_p1)#Positive Sample to Positive Data Pool
           Reference_Entropy 49.44715557808722
           Final_Entropy 49.447020249651644
Out[201]: 0.0001353284355758433
```

Fig: 2 DEM calculation and classification of data

# 4 Architecture

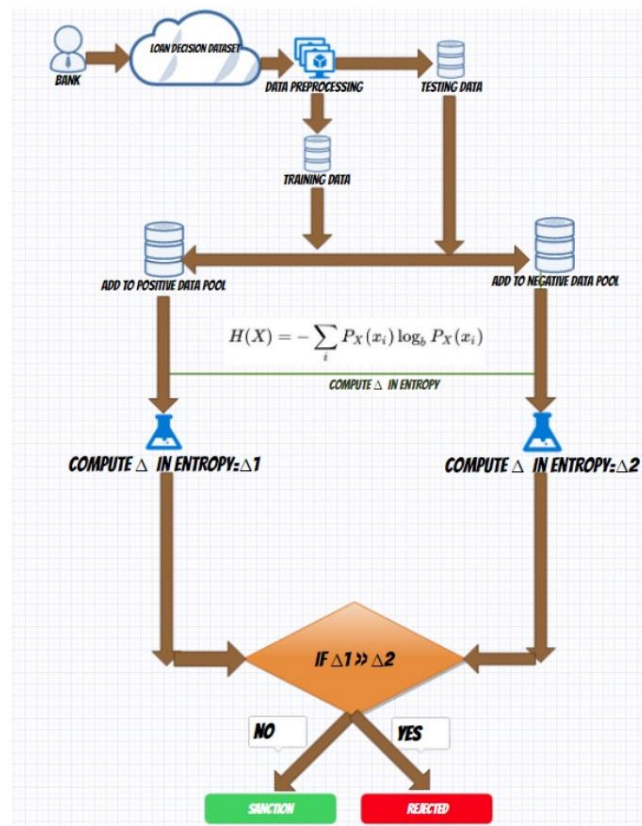

Figure 3: - Overview of procedure

# 5 Dataset

The dataset was obtained from https://github.com/avannaldas/Loan-Defaulter-Prediction-Machine-Learning. This dataset consists of total 5,32,428 cases, out of which 4,06,601 were defaulter cases and 1,25,827 cases were non-defaulter cases. The dataset consisted of total 45 columns. A precise and careful pre-processing of the dataset was performed. It contained more than 1,707,837 not defined or empty values out of 22,894,404 data entries and had to be carefully filled. Most importantly most of the data presented in the dataset was not categorical or absolute but had numerical values. The problem was to convert this data into categorical data, as entropy analysis of the attributes had to be done on categorical values. We took probabilities of occurrence frequency to calculate the attribute entropy of the dataset and for that we needed categorical values. So, the data had to be converted into categorical data. Certain parameters like "loan_status" were already in categorical binary form and needed nothing more than frequency analysis and mean filtering. The final and pre-processed dataset consisted of the following columns or attributes.

['member_id', 'loan_amnt', 'int_rate', 'pymnt_plan', 'dti', 'delinq_2yrs', 'mths_since_last_delinq', 'mths_since_last_record', 'open_acc', 'total_acc', 'mths_since_last_major_derog', 'application_type', 'purpose_car', 'purpose_credit_card', 'purpose_debt_consolidation', 'purpose_educational', 'purpose_home_improvement', 'purpose_house', 'purpose_major_purchase', 'purpose_medical', 'purpose_moving', 'purpose_other', 'purpose_renewable_energy', 'purpose_small_business', 'purpose_vacation', 'purpose_wedding', 'verification_status_Not Verified', 'verification_status_Source Verified', 'verification_status_Verified', 'grade_A', 'grade_B', 'grade_C', 'grade_D', 'grade_E', 'grade_F', 'grade_G', 'home_ownership_ANY', 'home_ownership_MORTGAGE', 'home_ownership_NONE', 'home_ownership_OTHER', 'home_ownership_OWN', 'home_ownership_RENT', 'final_desc', 'funded_amnt_B25_50', 'funded_amnt_B50_75', 'funded_amnt_G75', 'funded_amnt_L25', 'funded_amnt_inv_B25_50', 'funded_amnt_inv_B50_75', 'funded_amnt_inv_G75', 'funded_amnt_inv_L25', 'term_36 ', 'term_60 ', 'emp_length_ 1 ', 'emp_length_1 ', 'emp_length_10 ', 'emp_length_2 ', 'emp_length_3 ', 'emp_length_4 ', 'emp_length_5 ', 'emp_length_6 ', 'emp_length_7 ', 'emp_length_7 ', 'emp_length_8 ', 'emp_length_9 ', 'annual_inc_B25_50', 'annual_inc_B50_75', 'annual_inc_G75', 'annual_inc_L25', 'initial_list_status_f', 'initial_list_status_w', 'revol_bal_B25_50', 'revol_bal_B50_75', 'revol_bal_G75', 'revol_bal_L25', 'revol_util_B25_50', 'revol_util_L25', 'total_rec_int_B25_50', 'total_rec_int_B50_75', 'total_rec_int_G75', 'total_rec_int_L25', 'tot_cur_bal_B25_50', 'tot_cur_bal_B50_75', 'tot_cur_bal_G75', 'tot_cur_bal_L25', 'total_rev_hi_lim_G75', 'total_rev_hi_lim_L25', 'loan_status']

As it can be seen from the above list the final pre-processed dataset consisted of total 5,32, 428 rows and 88 columns. Certain columns like "sub_grade", "occupation_title" were dropped while making the final dataset. The original dataset consisted of a "description" column stating the applicant's reason for the loan application, certainly it consisted of long briefing paragraphs and had to be converted into categorical variables. To do so, we used a sentiment analysis model using the natural language processing library spacy and support vector linear classifier technique to analyze the sentiment of each description. This model was trained on a wide range of data and had an accuracy of 94.765%. Based on the classification of description column by the NLP model, we replaced the description paragraphs with respective sentiment analysis score (0 or 1) in the description column. Further, based on the distribution of numerical data in respective columns, the data was divided into the following four quantiles

Data >0.75 * (max value), 0.75*(max value) < Data > 0.50 * (max value), 0.50*(max value) < Data > 0.250 * (max value), Data< 0.25 * (max_value).

Based on these categories the data was converted into categorical values. The final dataset consisted of total 88 columns expanding from 41.

# 6    Result

The main motive of the project was to test and prove the reliability of the classifier that was proposed. The idea was to build a highly accurate classifier and test it against the conventional classifiers. Here, the size of the database had a mismatch. This mismatch was in terms of the size of batches of positive and negative cases.

      This mismatch wasn't an issue with our entropy-based classifier, as prediction was done on individual data points. But for our conventional classifiers, we had to look into the mismatches. The negative cases were 4 times the positive cases, hence 4 different datasets of equal size were used and calculated the accuracy of these conventional classifiers.

      The following two figures tell you about how our entropy-based classifier fares in comparison to the conventional classifiers:

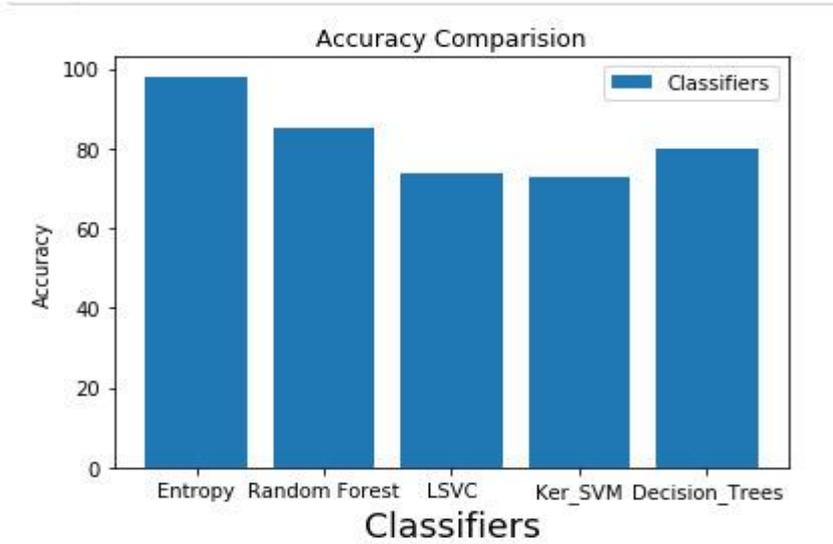

Figure 4: - The above graph is the comparison of accuracy of different classifiers

Table 1 :- Comparison f accuracies of various models

| Sr. no. | Classifier | Accuracy |
|---|---|---|
| 1. | **Entropy based classifier (The proposed classifier)** | **98.24%** |
| 2. | Random Forest Classifier | 82%-87% |
| 3. | SVM | 72%-74% |
| 4. | Kernel SVM | 71%-74% |
| 5. | Decision Trees | 77%-81% |

As discussed above, the entropy classifier is used for prediction, data point by data point. Hence each attribute's entropy is calculated for both positive and negative categories. The entropy behavior for both classes, column by column is presented in the below two figures. The x-axis has the attributes, while the y-axis has entropy magnitude:

Figure 5: - Values of entropy for all attributes (only positive cases)

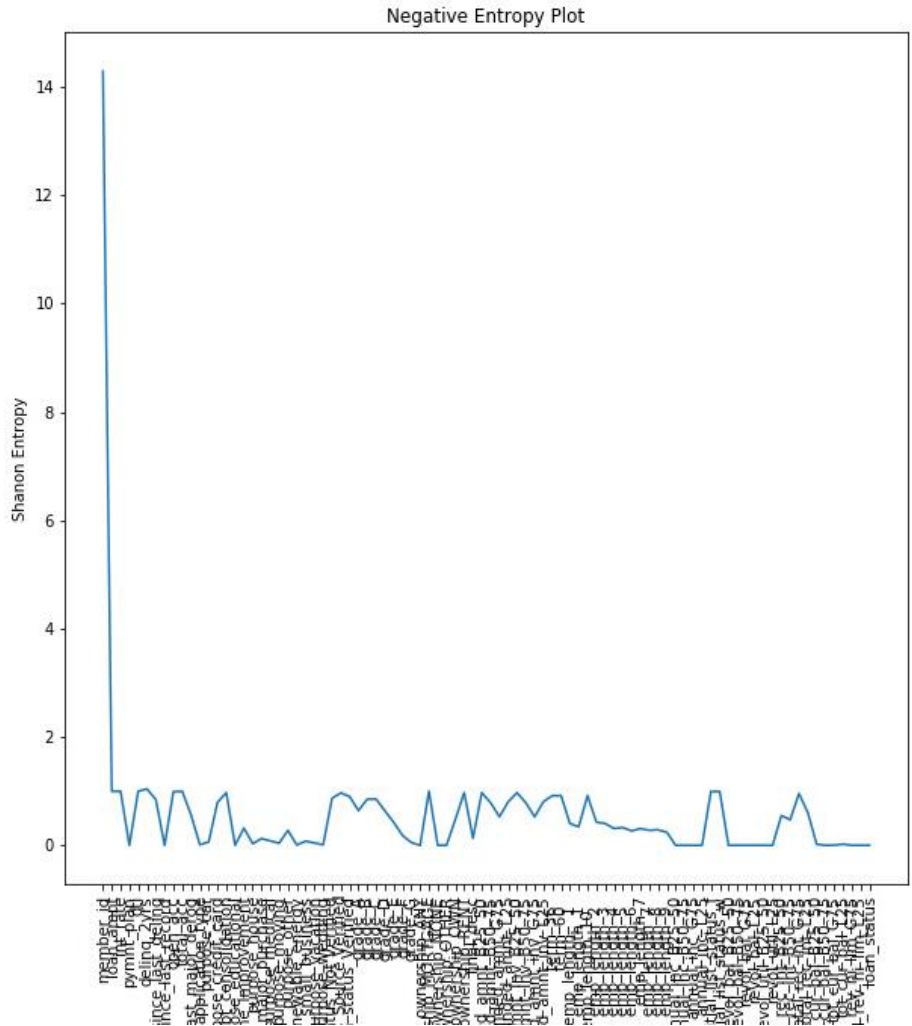

Figure 6: - Values of entropy for all attributes (only negative cases)

## 7 Conclusion

We discussed our strategy and methodology above. We then compared our classifier to the performance of conventional classifiers and it clearly suggested that the classifier we built, performed better in deciding which data points or entries will turn out to be NPA's. As suggested in our references, we also compared our model to the other models, built on similar lines. Apart from its soul major advantage of being highly accurate, we also understand the entropies of individual attributes, that tells us a lot about importance of each attribute which can be decisive in making decisions. So summing up, it's a better idea to use the entropy classifier, as the classifier as a whole gives a lot of important outputs, which are hugely accurate in determining the loan decision status and avoid all NPA losses.

# 8 References

[1] Entropy-Based Financial Asset PricingMiha´ly Ormos*, Da´vid ZibriczkyOrmos M, Zibriczky D (2014) Entropy Based Financial Asset Pricing. PLoS ONE 9(12): e115742 doi: 10.1371/journal.pone. 0115742

[2] Bawa, J. K., Goyal, V., Mitra, S. K., & Basu, S. (2019). An analysis of NPAs of Indian banks: Using a comprehensive framework of 31 financial ratios. IIMB Management Review, 31(1), 51-62.

[3] Pichler, A., & Schlotter, R. (2018). Entropy Based Risk Measures.

[4] Kadanda, D., & Raj, K. (2018). Non-performing assets (NPAs) and its determinants: a study of Indian public sector banks. Journal of Social and Economic Development, 20(2), 193-212.

[5] Arthi J., Akoramurthy B. (2018) Extrapolation and Visualization of NPA Using Feature Based Random Forest Algorithm in Indian Banks. In: Pattnaik P., Rautaray S., Das H., Nayak J. (eds) Progress in Computing, Analytics and Networking. Advances in Intelligent Systems and Computing, vol 710. Springer, Singapore

[6] Bardhan, S., Mukherjee, V. Bank-specific determinants of nonperforming assets of Indian banks. Int Econ Econ Policy 13, 483–498 (2016).

[7] Sudhakar, M., & Reddy, C. V. K. (2016). Two step credit risk assessment model for retail bank loan applications using Decision Tree data mining technique. International Journal of Advanced Research in Computer Engineering & Technology (IJARCET), 5(3), 705-718.

[8] Jayaraman, A. R., & Srinivasan, M. R. (2014). Performance Evaluation of Banks in India a Shannon-DEA Approach. Eurasian Journal of Business and Economics, 7(13), 51-68.

[9] Gang Wang, Jinxing Hao, Jian Ma, and Hongbing Jiang. 2011. A comparative assessment of ensemble learning for credit scoring. Expert Syst. Appl. 38, 1 (January 2011), 223–230.

[10] Alborzi, M., & Khanbabaei, M. (2016). Using data mining and neural networks techniques to propose a new hybrid customer behaviour analysis and credit scoring model in banking services based on a developed RFM analysis method. International Journal of Business Information Systems, 23(1), 1-22.